\documentclass[runningheads]{llncs}
\pdfoutput=1
\usepackage[T1]{fontenc}
\usepackage{multirow}

\usepackage{graphicx,verbatim, }
\usepackage{kotex, url} 
\usepackage{amsmath}

\usepackage{xcolor, colortbl}
\definecolor{cb_orange}{rgb}{1.0,0.51,0.0}
\definecolor{cb_green}{rgb}{0.3,0.67,0.29}
\definecolor{cb_red}{rgb}{0.89,0.1,0.11}
\definecolor{cb_pink}{rgb}{1, 0, 0.4}
\definecolor{cb_purple}{rgb}{0.5,0.0,0.5}
\definecolor{lightyellow}{rgb}{1, 1, 0.8}
\definecolor{dwkd_color}{rgb}{1, 0.93, 0.82}
\definecolor{grayrow}{gray}{0.9}

\usepackage{pifont}
\usepackage{xcolor}
\usepackage{booktabs}

\newcommand{\cmark}{\textcolor{green!80!black}{\ding{51}}}
\newcommand{\xmark}{\textcolor{red}{\ding{55}}}

\begin{document}
\title{LaGuadia: Language-Guided Adaptive Distillation from Pathology Foundation Models}
\titlerunning{LaGuadia: Language-Guided Adaptive Distillation from PFMs}
%
\author{Gangsu Kim \and Won-Ki Jeong$^\dagger$}
%
\authorrunning{Kim et al.}
%
\institute{Department of Computer Science and Engineering, College of Informatics, \\ Korea University, Republic of Korea \\
\email{\{gangsu1813, wkjeong\}@korea.ac.kr}}

  
\maketitle              

\def\thefootnote{$\dagger$}\footnotetext{Corresponding author: \email{wkjeong@korea.ac.kr}}

\begin{abstract}
Pathology Foundation Models (PFMs) offer powerful Whole Slide Image (WSI) representations but suffer from massive computational costs. 
While Knowledge Distillation (KD) can create efficient student models, existing multi-teacher methods often use suboptimal uniform weighting that ignores tissue heterogeneity. 
We propose LaGuadia (Language-Guided Adaptive DistillAtion), a framework that develops a compact pathology image encoder by dynamically integrating expertise from multiple PFMs under clinical linguistic guidance. 
Our approach utilizes a multi-stage pipeline: first, extracting visually observable clinical keywords from pathology reports; second, aligning visual features with these keywords via a Vision-Language meta-teacher (MedSigLIP) to provide dense semantic guidance; and finally, performing adaptive KD where teacher contributions are weighted based on their semantic alignment with the clinical narrative. 
Experiments on WSI captioning, visual question answering, and slide-level classification tasks demonstrate that an 87M parameter LaGuadia student model matches or exceeds foundation-scale models such as GigaPath and UNI, achieving strong factual consistency and robust generalization. These results highlight clinical language as an effective semantic anchor for building efficient and reliable digital pathology systems. Code is available at https://github.com/hvcl/LaGuadia.


\end{abstract}
\section{Introduction}
Computational pathology has rapidly advanced with the emergence of Pathology Foundation Models (PFMs), which leverage hundreds of millions of patches extracted from large-scale Whole Slide Image (WSI) datasets. Early studies primarily focused on vision-only PFMs that learn fine-grained morphological patterns in a self-supervised manner \cite{uni,phikonv2,gigapath,virchow2}. More recently, Vision-Language PFMs have further enhanced pathology representation learning by integrating visual information with clinical reports, either through joint embedding spaces \cite{conch} or fused architectures \cite{musk}. This integration enables richer diagnostic context beyond pure morphology, substantially improving representational expressiveness. However, the massive parameter scale of such PFMs incurs high computational costs, limiting their practicality in real-world clinical settings. Thus, developing lightweight yet expressive student models is crucial for improving the accessibility of digital pathology.

Knowledge distillation (KD) has emerged as an effective strategy for transferring the representational capacity of large PFMs to compact student networks. In pathology, KD methods can be broadly categorized into single-teacher \cite{h0mini,hviskd,rasa} and multi-teacher approaches \cite{gpfm,mtkd,pathryoshka,pathme}. While single-teacher distillation is constrained by the bias of a fixed expert, multi-teacher KD seeks to leverage complementary strengths across models. However, existing multi-teacher methods typically rely on uniform or coarse confidence-based aggregation, failing to reflect the strong spatial and semantic heterogeneity of pathology images. As a result, the optimal teacher may vary substantially across tissue regions.

Despite the need for adaptive teacher selection, assessing teacher expertise solely from visual features remains ambiguous due to the lack of a reliable ground truth, as visually similar tissue regions may correspond to distinct diagnostic interpretations depending on clinical context. 
Since pathological diagnosis is ultimately summarized in linguistic reports, we argue that such reports provide a more objective semantic reference for identifying clinically relevant knowledge. 
Although Rasa \cite{rasa} leverages report-derived linguistic information during distillation, it remains confined to a single-teacher model, limiting the diversity of transferable expertise.
Based on this insight, we propose a computationally efficient pathology image encoder trained via language-guided multi-teacher knowledge distillation, in which teacher contributions are dynamically weighted based on the semantic alignment between visual embeddings and pathological text. 
This context-aware expert selection enables more robust and generalizable representations across diverse pathology tasks.

In this work, we make several key contributions to the field of computational pathology. First, we introduce LaGuadia, a novel KD framework that adaptively integrates the specialized expertise of multiple PFMs by aligning visual features with clinical keywords in a shared semantic space, selecting teachers per patch via clinical language rather than the fixed or vision-only weighting of prior multi-teacher distillation, thereby resolving which teacher to trust for each tissue region under semantic heterogeneity.
Second, we demonstrate that a compact student model with only 87M parameters, fine-tuned with Low Rank Adaptation (LoRA) \cite{lora}, can achieve superior performance compared to PFM teachers such as GigaPath \cite{gigapath} (1.13B) and UNI \cite{uni} (303M) in complex generative tasks, including WSI captioning and Vision Question Answering (VQA). 
Finally, we provide extensive empirical evidence that our language-guided approach significantly enhances factual consistency in generated pathological narratives while maintaining robust generalizability across diverse slide-level diagnostic tasks, such as survival analysis and cancer subtyping.
\begin{figure}
\includegraphics[width=\linewidth]{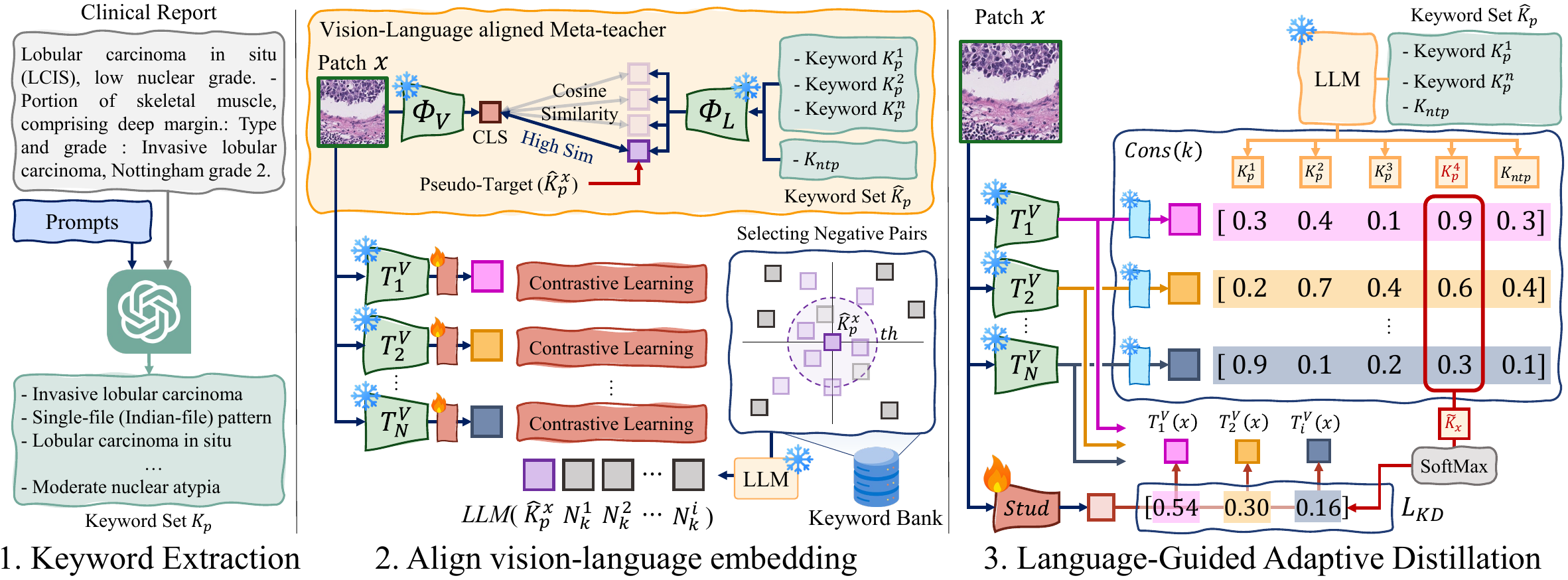}
\caption{Overview of the proposed framework comprises three stages.} 
\label{fig:overview}
\end{figure}
\section{Method}
Figure \ref{fig:overview} illustrates the overall framework of LaGuadia. 
The pipeline extracts clinically observable keywords, aligns visual embeddings through a vision–language meta-teacher, and adaptively distills teacher knowledge based on semantic agreement with the clinical narrative. The details of each stage are presented below. 
%
\subsection{Stage 1: Keyword Extraction} 
\label{method:stage1}
The first stage extracts clinical keywords from unstructured pathology reports to guide the image encoder's representation learning. To avoid non-visual noise—such as IHC or genetic alterations that are not visually observable in H\&E-stained images—we prioritize keywords with high visual-semantic correlation. We establish three pathological constraints: \textbf{(i)} observability on H\&E-stained WSIs, \textbf{(ii)} identifiability at 20x magnification, and \textbf{(iii)} minimal redundancy to maximize discriminative power in the semantic space. We employ GPT-5-mini~\cite{gpt5} as a controlled keyword extractor using pathology-specific prompts designed to favor visually observable concepts. Subsequently, the keywords extracted across all cohorts are aggregated to construct a Keyword Bank, which serves as a global semantic reference for negative samples during the contrastive learning process. 
%
\subsection{Stage 2: Align Vision-Language embeddings via Meta-Teacher}
While existing vision-only PFMs exhibit exceptional capabilities in representing morphological features, they lack a mechanism to directly link these features with clinical language. To address this limitation, we introduce a shallow projection layer integrated with the frozen visual backbones to perform an alignment process that incorporates linguistic knowledge into the visual representations.
\\ \\
\noindent\textbf{Patch-level Pseudo-labeling via Pathology Expert Meta-Teacher.}
\label{method:stage2}
To incorporate visual features with clinical semantic information in an environment where patch-level labels are unavailable, we perform weakly supervised learning via a Meta-Teacher $\Phi$. Specifically, we employ MedSigLIP \cite{medgemma} as our Meta-Teacher, leveraging its capabilities as a pathology-specific Vision-Language Model (VLM) to provide dense semantic guidance for the image patches.

A WSI contains diverse background regions that do not directly correspond to diagnostic keywords, in addition to the actual tumor areas. Consequently, it is unrealistic to assume that every patch within a WSI reflects the diagnostic characteristics described in the pathology report. To account for these domain-specific characteristics, we extend the patient-specific keyword set $K_p$ extracted in Section \ref{method:stage1} by appending a \underline{No Tumor Present} keyword $k_{ntp}$, representing non-diagnostic regions. This results in an expanded candidate set $\hat{K}_p = K_p \cup \{k_{ntp}\}$.

Given an input image patch $x$, the pseudo-label for the patch is defined as the keyword within the expanded set $\hat{K}_p$ of patient $p$ that has the highest cosine similarity in the VLM embedding space:
\begin{equation}
    \hat{K}^*_p = \arg\max_{k \in \hat{K}_p} \cos(\Phi_{V}(x), \Phi_{L}(k))
\end{equation}
Through this process, the model enhances semantic discriminability by aligning both tumor-specific morphologies and normal tissue patterns with their corresponding linguistic concepts.
\\ \\
\noindent\textbf{Similarity-based Contrastive Learning Strategy.}
To strengthen the alignment between the selected $\hat{k}_p^*$ and the visual embeddings, we perform contrastive learning using a SigLIP \cite{siglip} loss. To maximize the discriminability of the model, we dynamically reference $N$ negative samples from the Keyword Bank established in Section \ref{method:stage1}.

Rather than employing simple random sampling for the negative set, we propose an average similarity-based thresholding strategy to enable the model to effectively learn pathological differences. Specifically, we compute an average similarity $threshold$ by measuring the cosine similarity between the selected pseudo-label embedding and the patient-specific keyword embeddings. Keywords from the Keyword Bank whose embeddings fall below this threshold are selected as negative samples.
\begin{equation}
    threshold(\hat{K}^*_p, \hat{K}_p) = \text{avg}_{k \in \hat{K}_p} \left( \cos(\Phi_{L}(\hat{K}^*_p), \Phi_{L}(k)) \right)
\end{equation}
This strategy encourages the model to consistently repel concepts that are clearly distinct from the current patient's pathological distribution, thereby reinforcing the separation within the representation space. Furthermore, this serves as a critical foundation for enhancing the stability and accuracy of the language-guided weighting to be performed in Section \ref{method:stage3} 
\subsection{Stage 3: Language-Guided Adaptive Knowledge Distillation}
\label{method:stage3}
In the final stage, we perform adaptive knowledge distillation to develop a unified student model by integrating the specialized expertise of multiple teachers. The core of this process lies in determining adaptive weights that effectively incorporate the disparate representational strengths of each foundation model based on their clinical relevance.
\\ 
\noindent\textbf{Consensus-based Semantic Target Selection.}
To identify the core clinical concepts consistently highlighted by the ensemble of experts, we perform a soft voting process that aggregates similarity scores across the patient-specific keyword set $\hat{K}_p$. For each keyword $k \in \hat{K}_p$, which encompasses both diagnostic terms from the pathology report and the \underline{No Tumor Present} concept, we calculate the consensus score $Cons(k)$ as follows:
\begin{equation}
    Cons(k) = \sum_{i=1}^n \cos(Proj_i(T^V_i(x)), \Phi_L(k))
\end{equation}
The keyword with the highest soft voting score is selected as the Pseudo-Target Keyword ($\tilde{K}_{x}$) for the given patch. This process prevents knowledge conflicts between disparate models and establishes the most plausible clinical concept—as agreed upon by the ensemble of experts—as the anchor for knowledge transfer.
\\ \\
\noindent\textbf{Semantic Similarity Based Weighting.}
To quantify each teacher’s expertise, we compute the final distillation weights $\omega_i(x)$ by applying a Softmax function over the alignment scores, which represent the semantic proximity between the $i$-th teacher’s visual representation and the target $\tilde{K}_x$:
\begin{equation}
\omega_i(x) = \frac{\exp(\cos(Proj_i(T^V_i(x)), \Phi_L(\tilde{K}_{x})) / \tau)}{\sum_{j=1}^{n} \exp(\cos(Proj_j(T^V_j(x)), \Phi_L(\tilde{K}_{x})) / \tau)}
\end{equation}
This mechanism ensures that the teacher model most closely aligned with the clinical narrative receives a higher relative weight, allowing the student to adaptively distill knowledge from the most contextually relevant expert. 
\\ \\
\noindent\textbf{Adaptive Knowledge Distillation.}
Finally, the student model distills the class token representations from each teacher model based on the computed weights $\omega_i$. The distillation loss $\mathcal{L}_{KD}$ is designed to maximize the cosine similarity, encouraging the student model to integrally learn the morphological, contextual, and clinical expertise of the three expert teachers under the consistent guidance of clinical language.
\begin{equation}
    L_{KD} = \sum_{i=1}^n \omega_i(x) \cdot (1 - \cos(MLP_i(student(x)), T^V_i(x)))
\end{equation}
\section{Experiments and Results}
\subsection{Experimental Details}
\noindent\textbf{\underline{Datasets and Evaluation Protocol.}}
For a large-scale analysis, we extracted approximately 28 million patches from 1,910 WSIs and 1,801 corresponding pathology reports across three The Cancer Genome Atlas (TCGA) cohorts: BRCA, STAD, and THCA. To evaluate the generalizability of our learned representations, we conduct experiments across three diverse downstream tasks. For WSI Captioning, we employ the PathText dataset \cite{wsicaption}, and for VQA, we utilize WSI-VQA \cite{wsivqa} for BRCA and WSI-Bench \cite{wsillava} for STAD and THCA. Lastly, for the MIL task, we perform subtyping, survival, molecular, and progression analysis using the same TCGA cohorts. All experiments follow a slide-level 5-fold cross-validation protocol, and we report the mean performance across all folds. Notably, LaGuadia is self-supervised, using no downstream labels, and its Keyword Bank is built only from the train/val splits, preventing test-split leakage.
\\
\noindent\textbf{\underline{Evaluation Metrics.}}
We assess generated pathological narratives using METEOR (M) \cite{meteor}, BLEU-4 (B-4) \cite{bleu}, and ROUGE-L (R-L) \cite{rouge}. Furthermore, we adopt $Fact_{ent}$ \cite{fact} to assess the factual consistency and clinical reliability of the generated text, ensuring its alignment with ground-truth medical knowledge. For the MIL tasks, we utilize the C-Index for survival analysis and the weighted F-1 score for other classification tasks, such as cancer subtyping. In all tables, \textbf{bold} and \underline{underlined} values indicate the \textbf{best} and \underline{second-best} results, respectively.

\noindent\textbf{\underline{Implementation Details.}}
For preprocessing, CLAM \cite{clam} extracted valid tissue from WSIs, segmented into $224\times224$ patches at $20\times$ magnification. Our teacher ensemble comprises frozen backbones of GigaPath \cite{gigapath}, UNI \cite{uni}, and Virchow2 \cite{virchow2}. The student, a ViT-B (86M) initialized with DINOv3 \cite{dinov3}, was optimized using LoRA \cite{lora} on a single NVIDIA RTX A6000 (48GB). Stage 2 was trained for 10 epochs with a batch size of 4096 and 1 positive sample with 15 negative samples using AdamW with $lr=1e-3$. Stage 3 was optimized for 150k iterations with a batch size of 256. We applied AdamW with $lr=2e-4$ for LoRA and $lr=1e-3$ for the distillation MLP layer. 
For comparison, we benchmark our method against PFMs including GigaPath (1.13B), Virchow2 (631M), MedSigLIP (429M) \cite{medgemma}, UNI (303M), and MUSK (303M) \cite{musk}, as well as distillation baselines such as the single-teacher framework H0-mini (86M) \cite{h0mini} and the multi-teacher framework GPFM (303M) \cite{gpfm}.
\subsection{Results}
\noindent\textbf{\underline{WSI Captioning.}}
To assess the linguistic expressiveness of our visual representations, we replace the encoder in the HistoGPT \cite{histogpt} with various encoders. Table \ref{table:captioning} compares the generated captions against ground-truth on the PathText dataset. Despite its compact 87M parameters, our encoder achieves an overall score of 0.2461, comparable to far larger models like UNI (303M) and GigaPath (1.13B). Specifically, our model excels in the STAD and THCA cohorts. In STAD, it achieves the highest scores across all metrics, including METEOR (0.2896) and BLEU-4 (0.1031). For the THCA cohort, it records the top BLEU-4 score (0.1015), indicating that our features capture precise morphological details necessary for accurate description. These results show that LaGuadia matches billion-parameter models while remaining highly parameter-efficient.

\begin{table}[htbp]
\caption{Performance of WSI captioning on PathText benchmarks.} %
\label{table:captioning}
\centering
\resizebox{\columnwidth}{!}{
\fontsize{8.6pt}{10pt}\selectfont
\begin{tabular}{lcccccccccccc}
\hline
\multirow{2}{*}{Model} & \multicolumn{3}{c}{BRCA ($n=974$)} & \multicolumn{3}{c}{STAD ($n=316$)} & \multicolumn{3}{c}{THCA ($n=325$)} & \multirow{2}{*}{OVR} \\\cmidrule(lr){2-4}\cmidrule(lr){5-7}\cmidrule(lr){8-10}
          & M & B-4 & R-L & M & B-4 & R-L & M & B-4 & R-L & \\ \hline
GigaPath  & \textbf{0.2886} & \underline{0.0867} & \textbf{0.3575} & 0.2883 & 0.0973 & 0.3508 & 0.2755 & \underline{0.1008} & 0.3615 & 0.2452 \\ 
MUSK      & 0.2481 & 0.0636 & 0.3264 & 0.2620 & 0.0882 & 0.3272 & 0.2783 & 0.0985 & 0.3629 & 0.2284 \\ 
Virchow2  & 0.2801 & 0.0864 & \underline{0.3572} & 0.2857 & \underline{0.1004} & 0.3497 & 0.2745 & 0.0983 & 0.3654 & 0.2442 \\ 
MedSigLIP & 0.2703 & 0.0776 & 0.3417 & 0.2874 & 0.0986 & 0.3466 & 0.2765 & 0.1001 & 0.3602 & 0.2399 \\ 
UNI       & 0.2822 & 0.0856 & 0.3565 & 0.2887 & 0.0986 & \underline{0.3519} & \underline{0.2818} & 0.0996 & 0.3675 & \underline{0.2458} \\ \hline
GPFM      & 0.2794 & 0.0855 & 0.3562 & \underline{0.2891} & 0.0999 & 0.3502 & \textbf{0.2826} & 0.1003 & \textbf{0.3679} & 0.2457 \\ 
H0-mini   & \underline{0.2833} & \textbf{0.0872} & 0.3563 & 0.2850 & 0.0975 & 0.3498 & 0.2734 & 0.0981 & 0.3612 & 0.2435 \\ 
\rowcolor{dwkd_color}
Ours      & 0.2815 & 0.0852 & 0.3520 & \textbf{0.2896} & \textbf{0.1031} & \textbf{0.3524} & \underline{0.2818} & \textbf{0.1015} & \underline{0.3676} & \textbf{0.2461} \\\hline
\end{tabular}} 
\end{table}

\noindent\textbf{\underline{WSI VQA.}}
The visual encoder's capacity for complex clinical reasoning is evaluated on the WSI-VQA and WSI-Bench datasets by leveraging the WSI-VQA \cite{wsivqa}. This task requires the model to not only identify localized features but also to understand the spatial and relational context of the tissue.
As shown in Table \ref{table:vqa}, our encoder achieves the highest overall (OVR) score of 0.4184, albeit by a narrow margin over the second-best model (0.4182). This result demonstrates that the features extracted by our encoder possess superior reasoning capabilities across diverse pathological contexts. Notably, while showing a slight gap in linguistic fluency, our encoder yields significantly superior $Fact_{ent}$, reaching the highest scores in BRCA (0.9268) and STAD (0.4737). This divergence indicates that our language-guided adaptive distillation effectively prioritizes the alignment of visual features with precise pathological semantics, enabling the generation of reliable, fact-grounded responses.

\begin{table}[htbp]
\caption{Performance of WSI Vision Question Answering on WSI-VQA benchmarks.} 
\label{table:vqa}
\centering
\resizebox{\columnwidth}{!}{%
{\fontsize{8.6pt}{10pt}\selectfont
{\begin{tabular}{lcccccccccccccc}
\hline
\multirow{2}{*}{Model} & \multicolumn{3}{c}{BRCA ($n=8,599$)} & \multicolumn{3}{c}{STAD ($n=6,496$)} & \multicolumn{3}{c}{THCA ($n=6,427$)} & \multirow{2}{*}{OVR} \\ \cmidrule(lr){2-4}\cmidrule(lr){5-7}\cmidrule(lr){8-10} 
           &  M  & R-L & $Fact_{ent}$ & M  & R-L &$Fact_{ent}$ & M & R-L & $Fact_{ent}$ \\ \hline
GigaPath   & 0.2390 & 0.4709 & 0.9166 & 0.1714 & 0.3908 & 0.4700 & 0.1831 & 0.4067 & 0.4968 & 0.4161 \\ 
MUSK       & 0.2290 & 0.4608 & 0.9007 & \underline{0.1754} & \textbf{0.3970} & 0.4693 & 0.1827 & \textbf{0.4244} & \textbf{0.5067} & 0.4162 \\ 
Virchow2   & \underline{0.2445} & \textbf{0.4903} & 0.9086 & 0.1702 & 0.3846 & 0.4652 & 0.1812 & 0.4165 & 0.5027 & \underline{0.4182} \\ 
MedSigLIP  & 0.2208 & 0.4298 & 0.8913 & \textbf{0.1767} & \underline{0.3968} & \underline{0.4713} & 0.1822 & 0.4113 & 0.5004 & 0.4090 \\ 
UNI        & 0.2343 & 0.4626 & \underline{0.9178} & 0.1745 & 0.3912 & 0.4691 & 0.1782 & 0.4148 & 0.4926 & 0.4150 \\ \hline
GPFM       &\textbf{0.2448}& \underline{0.4770} & 0.9127 & 0.1747 & 0.3808 & 0.4710 & \underline{0.1846} & \underline{0.4191} & 0.4942 & 0.4176 \\ 
H0-mini    & 0.2358 & 0.4624 & 0.8890 & 0.1667 & 0.3822 & 0.4546 & 0.1809 & 0.4121 & 0.4942 & 0.4087 \\ 
\rowcolor{dwkd_color}
Ours       & 0.2373 & 0.4716 & \textbf{0.9268} & 0.1700 & 0.3848 & \textbf{0.4737} & \textbf{0.1855} & 0.4123 & \underline{0.5036} & \textbf{0.4184} \\ \hline
\end{tabular}}}}
\end{table}

\noindent\textbf{\underline{MIL Classification.}}
To evaluate representations in classification tasks, we adopt the ABMIL \cite{abmil} architecture within the MIL-LAB \cite{mil-lab}. Patch-level embeddings are aggregated to perform patient-level classification across multiple downstream tasks.
Table \ref{table:mil} summarizes the performance of our encoder on various MIL classification benchmarks. Despite being explicitly optimized for alignment with consensus-derived clinical keywords, our model exhibits strong generalization across conventional vision-based tasks, without any loss in performance. Notably, it achieves the highest overall (OVR) score of 0.6538, outperforming all compared pathology foundation models. In particular, our approach attains superior results on diagnostically demanding tasks, including LAUREN classification and N-stage prediction in STAD, as well as survival and progression in THCA. These results indicate that anchoring representations to clinically relevant semantics not only preserves but can even enhance the discriminative power of visual features in vision-centric downstream tasks.

\begin{table}[htbp]
\caption{Performance of MIL Classification on various benchmarks.} 
\label{table:mil}
\centering
\resizebox{0.94\columnwidth}{!}{%
\fontsize{8pt}{9.5pt}\selectfont
\begin{tabular}{lccccccccc}
\hline
\multirow{2}{*}{Model}  & \multicolumn{2}{c}{TCGA-BRCA} & \multicolumn{3}{c}{TCGA-STAD} & \multicolumn{2}{c}{TCGA-THCA} & \multirow{2}{*}{OVR} \\ \cmidrule(lr){2-3}\cmidrule(lr){4-6}\cmidrule(lr){7-8} 
                        &  Survival  & Subtype &  Survival  & LAUREN & N-Stage & Survival & Progression & \\ \hline
GigaPath  & 0.5467 & 0.8909 & 0.5119 & 0.7277 & \underline{0.2920} & 0.5162 & 0.8187 & 0.6149 \\ 
MUSK      & 0.5625 & 0.7585 & 0.4939 & 0.6382 & 0.1894 & 0.4974 & 0.8243 & 0.5663 \\ 
Virchow2  & \underline{0.5754} & \underline{0.8966} & 0.5122 & 0.7839 & 0.2379 & 0.5040 & 0.8216 & \underline{0.6201} \\ 
MedSigLIP & 0.5671 & \textbf{0.8974} & 0.5188 & 0.7107 & 0.2783 & 0.4307 & 0.8194 & 0.5974 \\ 
UNI       & 0.5536 & 0.7978 & 0.5320 & 0.8049 & 0.2469 & 0.5227 & 0.8261 & 0.6165 \\ \hline
GPFM      & 0.5184 & 0.8936 & \underline{0.5332} & 0.7599 & 0.2632 & \underline{0.5269} & 0.8347 & 0.6181 \\ 
H0-mini   & 0.5613 & 0.8890 & 0.4883 & \underline{0.8181} & 0.2603 & 0.4841 & \underline{0.8350} & 0.6199 \\ 
\rowcolor{dwkd_color}
Ours      & \textbf{0.5801} & 0.8930 & \textbf{0.5436} & \textbf{0.8303} & \textbf{0.3007} & \textbf{0.5843} & \textbf{0.8446} & \textbf{0.6538} \\\hline
\end{tabular}}
\end{table}

\noindent\textbf{\underline{Ablation Studies.}}
We conducted ablation studies on the THCA cohort to compare the baseline, uniform Knowledge Distillation (\textbf{KD}), and our Language-Guided (\textbf{LaGu}) mechanism. While naive KD marginally enhances surface-level linguistic metrics by mimicking descriptive patterns, it inadvertently leads to performance degradation in critical clinical benchmarks, such as $Fact_{ent}$ (0.5017 to 0.4988) and survival analysis (0.5425 to 0.5180), compared to the baseline. These results suggest that uniform teacher aggregation may not effectively differentiate between clinically relevant and less informative teacher signals, potentially leading to suboptimal knowledge transfer. As shown in Table \ref{table:ablation}, incorporating LaGu into the distillation process consistently outperforms naive KD, enhancing METEOR (0.2818) while significantly boosting $Fact_{ent}$ (0.5036) and survival analysis (0.5843). These results demonstrate that LaGu effectively balances linguistic fluency with diagnostic accuracy, grounding visual representations in clinical semantics.
\begin{table}[htbp]
\caption{Ablation study on the effect of Language-Guided Adaptive Distillation.}
\label{table:ablation}
\centering
\resizebox{0.85\columnwidth}{!}{%
{\fontsize{9pt}{10pt}\selectfont
\begin{tabular}{ccccccccccccccc}
\hline
\multicolumn{2}{c}{Method} & \multicolumn{3}{c}{WSI Captioning} & \multicolumn{3}{c}{WSI VQA} & \multicolumn{2}{c}{Classification} \\ \cmidrule(lr){3-5}\cmidrule(lr){6-8}\cmidrule(lr){9-10} 
KD & LaGu & M & B-4 & R-L & M & R-L & $Fact_{ent}$ & Survival & Progression \\ \hline
\xmark & \xmark & 0.2699 & 0.0943 & 0.3560 & 0.1834 & 0.4187 & 0.5017 & 0.5425 & 0.8122 \\ 
\cmark & \xmark & 0.2739 & 0.0983 & 0.3610 & \textbf{0.1856} & \textbf{0.4190} & 0.4988 & 0.5180 & 0.8105 \\ 
\cmark & \cmark & \textbf{0.2818} & \textbf{0.1015} & \textbf{0.3676} & 0.1855 & 0.4123 & \textbf{0.5036} & \textbf{0.5843} & \textbf{0.8446} \\ \hline
\end{tabular}}}
\end{table}

\noindent\textbf{\underline{Limitations/Future Work.}} We adopt class token-based distillation for simplicity and training stability. Extending the framework to patch-level supervision remains a promising direction for capturing finer morphological details. Finally, distilling per-fold rather than once across all folds could further reduce cross-fold leakage, which we leave as future work.

\section{Conclusion}
In summary, we introduced LaGuadia, a framework that adaptively distills expertise from multiple PFMs via vision-language alignment. By matching or exceeding foundation-scale models (e.g., GigaPath, UNI) in WSI captioning and VQA with only 87M parameters, we demonstrate that clinical linguistic guidance is a powerful surrogate for massive scale. Furthermore, the high factual consistency of our model highlights its potential as a reliable tool for digital pathology in resource-constrained settings.

\begin{credits}
\subsubsection{\ackname} This work was supported in part by the National Research Foundation of Korea under Grant RS-2024-00349697 and Grant RS-2021-NR060143; in part by the Institute for Information and Communications Technology Planning and Evaluation under Grant IITP-2026-RS-2020-II201819; in part by the Technology Development Program funded by the Ministry of SMEs and Startups (MSS) under Grant RS-2024-00437796; in part by the Commercialization Promotion Agency for R\&D Outcomes(COMPA) funded by the Ministry of Science and ICT(MSIT) under Grant RS-2026-25539491; and in part by Korea University Grant.

\subsubsection{\discintname} The authors have no competing interests to declare that are relevant to the content of this article.
\end{credits}

\bibliographystyle{splncs04}
\bibliography{Paper-6364}

\end{document}